# Constraints on the search space of arguments

Julio Lemos

Max Planck Institute for Comparative and International Law

University of São Paulo, Schools of Law (FDUSP) and Computer Science (IME-USP)

lazzarini.lemos@gmail.com

**Introduction**[1]

Arbitration proceedings frequently evolve into challenging disputes due to the complexity of the arguments brought forth by the parties and ultimately by the arbitrators issuing the award. For each claim *p* pro-arguments are presented by the claimant and con-arguments are presented by the defender.

According to some traditions accounting for valid reasoning taking place after the 'pleadings game', arbitrators should not look for reasons (grounds) outside this well defined 'search space' created by the parties. In certain civil law jurisdictions, the opposite holds. But the *communis opinio* holds that arbitrators also create arguments,[2] and "a well reasoned opinion can contribute greatly to the acceptance of the award by the parties by persuading them that the arbitrator understands the case and that his award is basically sound".[3]

Drawing from research on computational models of argumentation (particularly the Carneades Argumentation System), we explore the graphical representation of arguments in a dispute; then, comparing two different traditions on the limits of the justification of decisions, and devising an intermediate, semi-formal, model, we also show that it can shed light on the theory of dispute

---

[1] This paper was written thanks to a scholarship granted by the Center of Arbitration and Mediation of the Brazilian-Canadian Chamber of Commerce (CAM/CCBC) for a post-doc research stay at the Max Planck Institute for Comparative and International Law in Hamburg, Germany, during the second semester of 2013.
[2] G. B. Walker / S. E. Daniels, *Argument and dispute resolution systems*, in *Argumentation* 9 (1995), 698.
[3] F. Elkoru / E. A. Elkouri, *How arbitration works*, 4th ed., Washington, Bureau of National Affairs, 1985, p. 281.

resolution.

We conclude our paper with an observation on the usefulness of highly constrained reasoning for Online Dispute Resolution systems. Restricting the search space of arguments exclusively to reasons proposed by the parties (vetoing the introduction of new arguments by the human or artificial arbitrator) is the only way to introduce some kind of decidability -- together with foreseeability -- in the argumentation system.

## 1. Brief exposition of CMA and CAS

The field called Computational Models of Argumentation (CMA from here on) is now a mature research area, something that can be ascertained by acknowledging the success of the COMMA biannual conferences since 2006.[4] Having benefited from decades of research in a plethora of other fields such as philosophy, logic, law and computer science, CMA is the study of argumentation from the point of view of Artificial Intelligence.[5] The perspective which interest us here is that of law and theory of law, particularly dispute resolution, which sometimes dispense with non-trivial computational tools such as inference engines and could greatly benefit from visualization tools, as we will see. We will not try to survey the field, and neither will we propose a new immediate contribution to it, except as a guide to new practical research in the area and as an investigation of how CMA might contribute to the theory of dispute resolution and justification.

But how does CMA sees argumentation? Mainstream CMA could be understood by exploring the ASPIC framework, one of the most successful formal models of argumentation. The acronym stands for Argumentation Service Platform with Integrated Components; it was an initiative supported by the EU 6th Framework Programme whose goal was to "develop knowledge-based services based on semantically rich formalisms called Argumentation Systems".[6] We will instead quickly and informally outline Dung's abstract argumentation frameworks,[7] the core of ASPIC, so as to make things easier for the reader without a mathematical background. For a detailed investigation of ASPIC see the technical report by H. Prakken, one of its authors.[8]

---

[4] Visit http://www.comma-conf.org/.
[5] See T. Bench-Capon / H. Prakken, *Argumentation*, in A. R. Lodder / A. Oskamp (Eds.), *Information Technology and Lawyers: Advanced Technology in the Legal Domain, from Challenges to Daily Routine*, 2006, Springer, pp. 61–80; C. Reed / T. J. Norman, *A roadmap of research in argument and computation*, in same authors, *Argumentation Machines - New Frontiers in Argument and Computation*, Boston/Dordrecht/London, 2003, pp. 1-13.
[6] See the presentation at http://www.cossac.org/projects/aspic.

So here is how Dung's abstract argumentation framework looks like:

> An argument is a deduction whose premises are assumptions. To attack an argument, one attacks one or more of its assumptions. Thus an argument $a$ attacks an argument $b$ if and only if $a$ attacks an assumption in the set of assumptions on which $b$ is based. An argument attacks an assumption if and only if the conclusion of the argument is the contrary of the assumption. (The same goes for group attacks, from a set of assumptions to a set of assumptions.)
>
> In this context, an abstract argumentation framework (AF) is a pair $\{A, At\}$. Let $A$ be a set of arguments and $At \subseteq A \times A$ a relation of attack (argument $a$ attacks argument $b$) between elements of $A$. A set $B \subseteq A$ is called conflict-free if and only if there is no $a_i, a_j \in B$ such that $a_i$ attacks $a_j$. A set $B$ defends an argument $a_i$ if and only if for each argument $a_j \in A$, if $a_j$ dattacks $a_i$, then there exists an $a_k$ in $B$ such that $a_k$ attacks $a_j$. Informally, a set of arguments $B$ is *admissible* if and only if it contains only arguments that have not been attacked or that have been defended when attacked. The semantics is defined by the notion of a *preferred extension*. We won't delve into the details here.

In Dung's model, the internal structure of arguments does not matter (it is really an abstract model). We need a framework where it does matter, for that is what counts in dispute resolution systems. Thus, Thomas Gordon's Carneades Argumentation System (CAS, for short) is the most appropriate framework -- as we will see shortly -- since it was designed for use in a procedural context. Readers interested on the rich history of CAS might want to check out the related papers.

---

[7] P. M. Dung, O*n the acceptability of arguments and its fundamental role in nonmonotonic reasoning, logic programming and n-person games*, in *Artificial Intelligence* 77, 2 (1995), 321-357.
[8] *An abstract framework for argumentation with structured arguments*, technical report, University of Utrecht, 2009.

To lay down the CAS, basically we need only define statements, premises, arguments and argument graphs:

> Definition 1. (Statement) A *statement* (*s*) is a declarative sentence in a language. The complement of a statement *s* is denoted by -*s*.

> Definition 2. (Premise) A *premise* (*p*) is a statement that falls on one of the following categories, and nothing more: an *ordinary premise* (*so*), an *assumption* (*sa*) or an *exception* (*se*). *P* is the set of premises.

> Definition 3. (Argument) An argument is a tuple $<c, d, p>$ where *c* is a statement, *d* is a member of the set $\{pro, con\}$ and $p \in 2^P$. Note that *c* denotes the conclusion, *d* the direction (*pro* or *con*), and *p* the premises of the argument which may be obtained by applying the correspondent functions in this way: *c(a)*, *d(a)*, *p(a)*.

> Definition 4. (Argument graph) An *argument graph* is an acyclic, labeled, finite, directed, bipartite graph consisting of argument nodes and statement nodes. Edges link up premises and conclusions in the arguments. At most one statement node is allowed for *s*, -*s*.

Argument graphs represent justifications, so that there is an acceptability relation between an argument graph and statements, with the whole framework aimed at modelling proofs. A statement, which may be used in several arguments, is thus *acceptable* if and only if the corresponding argument graph is a proof thereof. In this system, a dispute is about statements *s*, -*s*, that is, about the final acceptability of *s* or of its complement (if *s* is accepted, then -*s* is rejected and vice-versa).

The advantage of argument graphs is that they can be easily visualized. In figure 1 below, we give an example of an argument graph for a fragment of a dispute about the applicability of the Convention on the International Sale of Goods (CISG). The main issue is *s* (CISG applies) or -*s* (CISG does not apply):

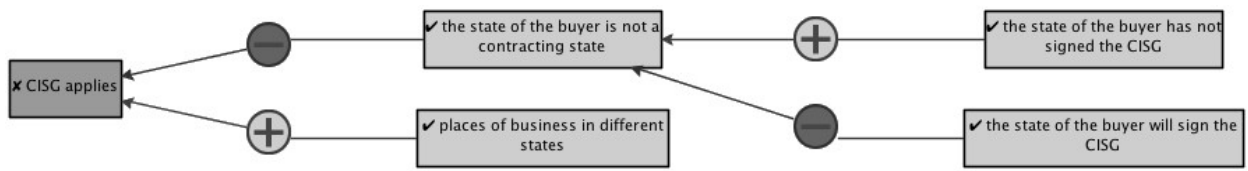

The case we chose to illustrate the CAS is simple. According to Article 1(1) of the CISG, the convention applies to contracts of sale of goods between parties whose places of business are in different states. Article 1(1)(a) presents a further premise: both states need to be contracting states. That means both places of business (of each of the parties) must be in contracting states, a premise that depends on the fact that the state has already signed the convention. In our toy case, we have the following fictional facts, some of which are not directly depicted in the graph: 1) seller and buyer have places of business in different states (statement *s1*); 2) the state where the buyer has settled his business is England, and England is has not signed the convention (*s2*); 3) England has announced that it will sign the convention (*s3*); 4) from *s2*, we conclude that the state where the buyer has his place of business is not a contracting state (*s4*). In the graph, arguments are denoted by circles, and statements by boxes. The use of arrows is self-evident: arguments go from premises (ordinary premises, assumptions or exceptions) to a conclusion. The "+" sign denotes a *pro* argument, and the - sign denotes a *con* argument. So we establish the following acceptability relations: all statements above are accepted. The fact that *s3* was put forth by the claimant (say) as a desperate *con* argument against *s4* does not prevent us from accepting *s4* (due to the internal structure of the dispute). The conclusion *c*, "CISG applies", is therefore negated: the *pro* argument *s1* is clearly weaker than the *con* argument *s4*.[9] Why so? Article 1(1)/1(1)(a) provides that the CISG itself applies to contracts of sale of good between parties whose places of business are in different states, *except* when one or both of these states are non-contracting states (or, alternatively, *provided* that both these states are contracting states). The *x* before the statement "CISG applies" denotes that the argument has been rejected, as expected, for if the premises of a *con* argument hold, then we have sufficient reasons for rejecting the conclusion. It is a very simple example which could otherwise be a very complicated one, with premises attacking assumptions attacking premises undermining conclusions and so on. Regardless of complexity, virtually any case could be represented isomorphically at the argumentation level using CAS.[10]

---

[9] For inferential purposes, in CAS arguments can be given degrees of strength by using real numbers in the interval [0, 1].

[10] For a detailed representation of a very complex and well known case in US litigation, see T. F. Gordon /

The brief exposition and example above illustrate the use of argument graphs for representing the network of statements brought forth by the parties and the arbitrators as a way of consistently arguing about the case or justifying a decision. Although overlooked or ignored by the legal community, it is a very powerful tool (or tools: from simple visualization to sophisticated inference engines) both for solving practical cases and for investigating difficult theoretical issues in the fields of legal theory and dispute resolution systems.

Having said that, we now turn on to the problem of justification itself.

## 2. Two accounts on the search space of legal argumentation (*inventio* functions)

A search space or solution space in mathematics and computer science is the set of all possible solutions to a given problem. In chess, for instance, the search space is the set of all valid moves, given a game state. Each point in the corresponding search state represents a possible solution to the problem; usually one deals not with arbitrary solutions, but with feasible ones.

Analogically, we could think of the possible solutions to a dispute as the search space of argumentation.[11] The example given above illustrates that. The case is: does the CISG apply? There are two main solutions: either it applies (*s*) or it does not (-*s*). Arguments pro and con each of the main solutions, together with the resulting statement are possible complete solutions. They are virtually infinite in general terms, but finite *given a case* and a record of the arguments effectively used in the pleadings game: (*s*\*1) CISG applies because the parties have places of business in different states; (*s*\*2) CISG applies because the parties have places of business in different states and it has not been proven that the state where the buyer has his place of business is a non-contracting state; (-*s*\*1) CISG does not apply because the parties have places of business in different states but it has been proven that the state where the buyer

---

D. Walton, *A Carneades representation of Popov v Hayashi*, in *Artificial Intelligence and Law* 20, 1 (2012), 37-56.

[11] L. T. McCarty's final remarks on a AI & Law paper by Kevin D. Ashley are worth mentioning: "Let's face it: The Law is AI-Complete! This means, by analogy with the theory of NP-Completeness [...] that given any problem *X* in AI there exists a polynomial-time mapping from *X* into some problem *Y* in a particular legal domain. In order to make progress in this field, we need to identify tractable problems within AI whose solutions, when applied to the law, yield either genuine theoretical understanding or real practical applications, or both." (T. Bench-Capon et al., *A history of AI and Law in 50 papers: 25 years of the international conference on AI and Law*, in *Artificial Intelligence and Law* 20, 3 (2012), 13.1).

has his place of business is a non-contracting state; and so on (there are virtually infinite solutions!). But suppose the arguments $s^*1$, $s^*2$ and $-s^*1$ above are either the only *selected* solutions. Then {$s^*1$, $s^*2$, $-s^*1$} represents the effective search space of the case. Each of these statements, provided they are accepted, represent a point in the search space.

We call here *inventio* the process of systematically looking for (plausible) arguments pro and con a given conclusion, i. e., the solution to a case. It is not without reason. The term is exactly the one used by Ancient writers to refer to the core of the art of argumentation. It is one of the five canons of rhetoric (*disposition, memoria, pronuntiatio*, *elocutio*, *inventio*), denoting the systematic search for arguments. As put by Cicero himself: *inventio est excogitatio rerum verarum aut veri similium, quae causam probabilem reddant*.[12] (A fancy alternative would be to explore the expression *ars combinatoria*, used by Leibniz in his early career to denote not only mathematical operations (roughly what is done today under the name of Combinatorics), but also the general ability to explore the search space of a legal case.)

*Inventio*, for short, is the obtaining of all interesting points in the search space of a given case.

Here is where classical rhetoric and theory of law meet. There are two traditions informally accounting for the constraints on the size of the search space of any given case. Although it is not exactly a good name in a precise historical context, we will call them the "civil law" (CiL) and the "common law" (CoL) accounts. Civil law, on the one hand, has it that the judge should be given only the facts of the case so as to be in a position to reach a decision; even though the parties will provide him or her with a plethora of arguments, the issue of legal justification is the judge's alone to solve. *Da mihi facti, dabo tibi ius*. On the other hand, common law has it that the judge is to restrict his *inventio* (the search for arguments) to the arguments expressly provided by the parties. He is not allowed to go beyond those arguments.

Now we explore some formal properties in a semi-formal way. It will be useful for understanding what is going on in the intended domain, even when some important elements are left undefined.

*Inv* is a function which takes as input a case $\gamma$ and returns the set of all points in the search space $S$ of $\gamma$. Let $\gamma$ be a case. Then $inv(\gamma) = S = \{s_1, s_2, ..., s_i, -s_1, -s_2, ..., -s_j\}$ for $i, j \in \mathbb{Z}^+$, $i = j$, since in our domain of discourse for every $s$ there is an $-s$ obtained by negating $s$ (*complement* may be seen as a function mapping a statement to its logical complement). A solution is a statement node in an argument graph $D$ (a member of the set of all possible argument graphs for a case) supported by an argument node $a$ also in $D$. Recall that arguments are represented in CAS as a tuple $<c, d, p>$ where $c$ is a statement, $d$ is a member of the set $\{pro, con\}$ and $p \in 2^P$. A further condition may be imposed on the solution, to wit, that it is *accepted*. A solution being accepted means that the argument graph is a proof of the corresponding statement; and so each of the different solutions in $S$ induce different argument graphs. The fact that argument graphs are cyclic ensures decidability for a given a solution.

---

[12] *De inv.* I, 9.

Since (1) $S$ tends to be virtually infinite ($i$ and $j$ may be arbitrarily large) and not even recursively enumerable, and (2) our intended domain is not fitted for infinite outputs, we need a way of constraining its size. An adequate *inventio* function $inv^*$ is similar to an *inventio* function $inv$ except that given a case $\gamma$ it returns only *adequate* points, which must be finite in number. An adequate point set $P \subseteq S$ is simply the result of applying a selection function to the output search space $S$ of $inv(\gamma)$. The adequacy of a point is given by a system of justification imposing the corresponding selection function. And that is where different accounts on adequate justifications may be of help.

In our semi-formal framework, two extreme solutions come to mind, recalling what we said about CiL and CoL. The first is that the selection function should return only solutions $P \subseteq S$ proposed by the parties in the pleadings stage (only arguments effectively brought forth); the second is that the selection function should return some or all of the solutions proposed by the parties $P \subseteq S$ and some new solutions in a possibly empty subset $R \subseteq S$, so that the output is their union $P \cup R \subseteq S$. Either way now the proper subset $P$ of $S$ is finite and well-defined and can (given an entirely transparent, formal, framework) be searched over by an algorithm with the proper questions, such as "give me all the solutions using at least one Argument from Expert Witness". $S$ cannot be searched because it is not even recursively enumerable, even given a normative system, principles, goals, cases, and all the facts that can be seen as established. The problem is that $R$ has the same practical disadvantage of $S$ itself: it is possibly infinite. While we have a source for $P$, the pleading stage with its dialectical constraints, the possible elements of *R* are virtually the same as those of *S*. Recall Solomon's veterotestamentary decision in 1 Kings 3:16-28: given solutions by mother_1 and mother_2 about who was the mother of the living child (both elements of *P* in our artificial framework) he rejected both arguments and decided to divide the child in two (therefore, in our framework, an element of *R* entirely outside the search space of *S*). Of course it was only an strategic solution. Such Solomonic solutions are not uncommon in arbitrations and have even been given a name: "splitting the baby".

We introduce the notion of argumentative agents for representing the parties and the form *someone* a *proposed a solution* s *in the pleadings game*, provided that it was done in a valid way,[13] formalized as *prop(a,s)*. Since the pleadings game is always implicit, we omit it from the formalization.

The intended function *inv\** (the first case above) is relatively trivial and can be defined as follows:

> (Definition of *inv\**) Let $\gamma$ be a case and $inv(\gamma) = S = \{s_1, s_2, ..., s_i, -s_1, -s_2, ..., -s_j\}$ for $i, j \in \mathbb{Z}^+$, $i = j$. Let $\alpha$ and $\beta$ be the only argumentative agents in the pleadings game. Then $inv * (inv(\gamma)) = P = \{s_1, s_2, ..., s_m, -s_1, -s_2, ..., -s_n\}$ for $i, j, m, n \in \mathbb{Z}^+$ such that $i = j$, $m < i$, $n < j$ and for every $s_k \in P$, either $prop(\alpha, s_k)$ or $prop(\beta, s_k)$.

Since the second extreme case possibly includes *R*, we only need a function *inv\*\** that combines *inv* and *inv\**. That can be easily done, but we will explore a more interesting intermediate case, that seems to reflect empirical data on the justification of awards in arbitration. First we have to understand what are the informal constraints on *inventio* in arbitration proceedings.

## 3. Theoretical and empirical note on *inventio* and justification in arbitration

This paper is based on two common sense assumptions: (a) all international arbitration systems of note require that at least final decisions be justified by precisely indicating the reasons upon which they are based;[14] (b) implicitly or otherwise, the majority of systems also provide for that awards be based only on the matters put before the arbitrators by the parties; consequently, that the award shall not deal with arguments that were not discussed during the arbitration.[15]

---

[13] Recall the notion of an *argument move* discussed in the literature, but note that it is a final move and a complete solution (a statement backed by an argument) in an argument graph.
[14] See, for instance: UNCITRAL Model Rule, Article 31(2); ICC Rules of Arbitration 2012, Article 31(2); LCIA Rules 26.1; AAA International Dispute Resolution Rules, 27.2; ICSID Convention, Article 48(3). Some systems provide that the parties may agree otherwise. See Ph. Fouchard / E. Gaillard / B. Goldman, *Traité*

Ideally, a good decision is not one for which there are some good reasons for it, but one for which one could say that "on balance, all reasons considered, the decision represents the best resolution of the available reasons."[16] Of course, we understand the already mentioned problem raised by the computer scientist Thomas F. Gordon: it is impossible to consider *all reasons*, since the set of *possible arguments* is not even recursively enumerable.[17] That is why we think the state of the otherwise dynamic set of arguments presented by the parties during the pleadings stage is the mandatory starting point for the effective argumentation space of justification in arbitration. Perhaps only argumentation theory and CMA could convey this idea.

The goal of an award being not only to present a good solution, but also -- and perhaps more importantly -- to persuade the parties that it is a good solution, it is clear that the investigation on the bounds of *inventio* is of utmost importance.

Consider the usual, canonical, form of justification in arbitration (and other dispute resolution systems, including judicial rulings): as regards item $n$ of the dispute, $\alpha$ claimed that $x$ because of $X$ and $\beta$ claimed that $y$ because of $Y$; the arbitrator decides that $z$ because of $Z$. (Notice that $X$, $Y$ and $Z$ are nonempty sets of grounds in the intended domain, provided that one accepts the assumption (a) above.) One can easily see that the following bounds on $z$ and $Z$ are the only ones that may be logically conceived of: (1) the arbitrator may decide that $z$ because of $Z$ if and only if either $z = x$ and $Z = X$, or $z = y$ and $Z = Y$; (2) the arbitrator may decide that $z$ because of $Z$ if and only if either $z = x$ or $z = y$; (3) the arbitrator may decide that $z$ because of $Z$ if and only if either $Z = X$ or $Z = Y$; (4) the arbitrator may decide that $z$ because of $Z$ regardless of $x$, $y$ and

---

*de l'arbitrage commercial international*, Litec, 1996. This is not, and notably has not been, always true in Common Law systems and in some English speaking countries, although the most recent trend strongly favors reasoned awards; see R. David, *Arbitration in international trade*, Kluwer, 1985, pp. 319-320.

[15] L. A. Mistelis (ed.), *Concise international arbitration*, Kluwer, 2010, p. 364. The most immediate reason for that kind of provision is -- regarding the worst case scenario -- that awards deciding *ultra* and *citra petita* may might not be recognized by states that signed the New York Convention, with Article V(1)(c) being possibly invoked.

[16] R. Sartorius, The justification of the judicial decision, in *Ethics* 78, 3 (1968), 178. That is not the opinion of Sartorius, but a depiction of a common theory on justification.

[17] *Foundations of argumentation technology - summary of habilitation thesis*, Technische Universität Berlin, 2009, p. 21.

grounds thereto *X*, *Y*. Otherwise (and less precisely) stated,[18] the arbitrator's argumentation may be bound both by the premises and the conclusion that immediately supports the claim as presented by the parties to the dispute; only by the premises; only by the conclusion; or by neither of them.

Before we proceed with the investigation in the next section, we briefly summarize theoretical and empirical findings related to the issue above.

On the one hand, the theoretical status of this question is dubious. Commentators rarely address the issue with the required precision, perhaps due to their focus on non-formal approaches. We do not think that this superficiality is wise; that does not even favor pragmatism, for superficiality leave all the practical problems entirely open. Likewise, philosophers of law either address the problem of the search space of arguments in arbitration only indirectly, for local law codes in the CiL tradition and CoL provide different and mutually inconsistent answers, or they simply ignore it.[19] The most important problem, in our view, can be reduced to a single twofold question: is it permitted for arbitrators to fabricate one or more arguments that were not even mentioned, or implied, by the parties to the dispute, and if so, do these arguments include only conclusions or also premises?

On the other hand, in practice, approaches vary greatly. A certain empirical analysis in the field of jurimetrics could, notwithstanding, illuminate the issue. It was conducted by Ole Kristian Fauchald of the University of Oslo; a detailed investigation of motivation (among other problems) using a *corpus* of nearly 100 arbitration awards issued by *ad hoc* tribunals of the International Centre for the Settlement of Investment Disputes (ICSID) revealed that they "made an effort to address all arguments raised by the parties to the dispute", but that "this does not prevent[ed] tribunals from exercising judicial restraint by avoiding dealing with issues that can be left aside as a consequence of conclusions on other issues"; particularly, "tribunals responded explicitly and in detail to all arguments of the parties to the dispute".[20] It seems that in the majority of decisions, arbitrators used the arguments presented by the parties as the starting point for the work of justifying their conclusions, but that they did not consider the state of arguments to be a necessary bound on their discretion. Our own brief and *ad hoc* quantitative research on the *corpus* of 2001-2007 awards issued by arbitral tribunals under the ICC Rules and administration [21] allowed us to reach the same conclusion. Most notably: (1) sometimes both parties used the same argument, and the arbitrators concurred, adopting exclusively the premises and the conclusion presented by them;[22] (2) sometimes the arbitrators raised an issue not discussed by the parties and fabricated an entirely new argument alleging it was a public law issue, but following a mutual agreement by the parties;[23] (3) sometimes the arbitrators considered premises and conclusion presented by the claimant and premises and conclusion presented by the defendant and decided by adopting either the whole argument of the claimant or the whole argument of the defendant;[24] (4) on one occasion the parties agreed upon an issue, presenting their arguments, and the tribunal, noting the agreement, simply adopted the consensual interpretation, without presenting its own arguments;[25] (5) on one occasion, the tribunal considered the argument of the parties, adopted the conclusion of the claimant but, without properly rejecting his premises, adopted a different premise;[26] (6) on the same occasion, the tribunal adopted the premise presented by the defendant, which was the same as the one presented by the claimant, but rejected the conclusion of the former, concurring with the (different) conclusion of the latter.[27]

---

[18] This kind of imprecise restatement of formal or semi-formal propositions is dangerous but can be used to facilitate understanding of the main issue.

[19] For some general discussions on the justification of decisions, See I. Scheffler, *On justification and commitment*, in *Journal of Philosophy* 51 (1954), 180-190; N. MacCormick, *The artificial reason and judgement of law*, in *Beheft* 2 (1981), 105-120; A. Arnio / R. Alexy / A. Peczinik, *The foundation of legal reasoning*, in *Rechtstheorie* 12 (1981), 423-448; R. Sartorius, *The justification of judicial decision*, in *Ethics* 78 (1968), 171-187; A. Peczenik, *Grundlagen der juristischen Argumentation*, Springer, 1983.

The findings allow us to conclude, further, on plausible theoretical, legal and empirical grounds that unless the case deals with a public issue, the tribunal is *defeasibly* allowed to vary on the sets of grounds *X, Y* presented by the parties, but not on the set of claims raised by the parties *x, y*. As regards public issues, it seems reasonable that they are allowed to be raised by the arbitrators; but in that case it should be decided only after consultation with the parties, since one of the main goals of the award is to present a rhetorically correct solution to the case.

---

[20] *The legal reasoning of ICSID tribunals - an empirical analysis*, in *European Journal of International Law* 19, 2 (2008), 315.
[21] *Collection of ICC arbitral awards 2001-2007*, Wolters Kluwer, 2009.
[22] For instance, case no. 7645 of 1995, published in 11 ICC Bulletin 2 (2000), 34-46.
[23] For instance, case no. 8423 of 1994, unpublished.
[24] For instance, case no. 8445 of 1996, unpublished.
[25] Case no. 8790 of 2000, unpublished.
[26] Case no. 10274 of 1999, unpublished.
[27] Idem.

## 4. A possible formalization of the constraints. Conclusion

Based on our conclusions above, it seems that the best way of attacking the issue of the constraints on the search space of argumentation in arbitration, aiming at proposing a 'persuasion-driven', rhetorically correct model of award drafting, is to follow an algorithmic approach.

The parties to a dispute, at the time of the decision, have already presented their cases. All the arguments were brought forth, and the arbitrator is now bound both by the primary duty to decide (*non liquet* not being a possible way out) and by the duty to decide within the limits drawn by the issues and arguments presented. The duty is one of invention. Recall the CAS sketched above and the functions *inv\**, *inv* defined. Given the result of the *inventio* already carried forth, in a dialectical way, by the parties, the arbitrator should then apply his own *inventio* to the result; that should bring us back to the problem sketched in the last section, which has four basic possible solutions: should his *inventio* be bound both by the premises and the conclusion presented by the parties to the dispute, only by the premises, only by the conclusion, or by neither of them? Based on the conclusions established in last section, he is bound only by the conclusions *x* or *y*, which may be identical, but not by the sets of grounds *X*, *Y*. For each item *n* of the dispositive section, he may use a new set of grounds *Z* to back his decision in favor of conclusion *x* or conclusion *y*, even when faced with a public issue (that should be added to the premises already in *Z*). But what is the use of *X* and *Y*? Utterly ignoring them would be a violation of the *purpose* of basic due process principles such as that of contradiction (*audiatur et altera pars*). It would be absurd to hear both parties if their arguments should in be ignored after all.

That have been said, we are ready to present our model. Let CAS be the underlying argumentation system, with working definitions. Let $\gamma$ be a case with dispositive items $\{n_1, n_2, ..., n_i\}$ and $inv(\gamma) = S = \{s_1, s_2, ..., s_i, -s_1, -s_2, ..., -s_j\}$ for $i, j \in \mathbb{Z}^+$, $i = j$. Let $\alpha$ and $\beta$ be the only argumentative agents in the pleadings game. Let $\Gamma_k \succ s_k$ be a solution *s* backed by a set of reasons. The set of party-solutions (or search space of argumentation as defined by the parties) is then:

$$P = \{\Gamma_1 \succ s_1, \Gamma_2 \succ s_2, ..., \Gamma_m \succ s_m, \Delta_1 \succ -s_1, \Delta_2 \succ -s_2, ..., \Delta_n \succ -s_n\} \text{ for } m, n \in \mathbb{Z}^+ \text{ such}$$

that for every $\Gamma_k \succ s_k \in P$, either $prop(\alpha, \Gamma_k \succ s_k)$ or $prop(\beta, \Gamma_k \succ s_k)$.

The *inventio* function representing the selective activity of an arbitrator given the search space of argumentation *P* provided by the parties can be thus characterized as follows:

> *inventio*(*P*) = for each dispositive item *n* in the decision, nondeterministically either (a) $P/prop(\alpha, \Gamma_k \succ s_k) \cup S$ such that for each $s_l \in S$, $s_l = s_k \in P$ and $prop(\beta, s_l)$ or (b) $P/prop(\beta, \Gamma_k \succ s_k) \cup S$ such that for each $s_l \in S$, $s_l = s_k \in P$ and $prop(\alpha, s_l)$.

That means, for each item of the decision, either the *inventio* results in a solution proposed by $\alpha$ together with reasons invoked by $\alpha$ in addition to new reasons invoked by the arbitrator; or in a solution proposed by $\beta$ together with reasons invoked by $\beta$ in addition to new reasons invoked by the arbitrator.

A conservative approach -- sometimes found in practice, as our investigation above reveals -- would be to further constrain *inventio*(*P*), vetoing new reasons (premises) introduced by the arbitrator. That would demand a different *inventio* function defined as follows:

> *inventio**(*P*) = for each dispositive item *n* in the decision, nondeterministically either (a) $P/prop(\alpha, \Gamma_k \succ s_k) \cup S$ such that for each $s_l \in S$, $s_l = s_k \in P$ and $prop(\beta, \Gamma_k \succ s_k)$ or (b) $P/prop(\beta, \Gamma_k \succ s_k) \cup S$ such that for each $s_l \in S$, $s_l = s_k \in P$ and $prop(\alpha, \Gamma_k \succ s_k)$.

This second approach is likely to simplify the formal apparatus, facilitating computational implementations of argumentation models, but may be seen as unrealistic in the light of the real-life functioning of the majority of human-centered systems.

We note -- most importantly -- that Online Dispute Resolution (ODR) could favor this second approach (using the strongly constrained *inventio** to obtain the search space of argumentation), since the enlarging of the search space is computationally unfeasible, in addition to foster total undecidability (solutions are not recursively enumerable). It is also impossible to automatically generate new premises, unless a new apparatus is introduced for that purpose. Nonetheless, it is doubtful that such an ODR system would be unfair.[28] Constraints on the search space of argumentation can be justified by showing that foreseeability is to be favored over traditional material justice. In that way it is possible to transcend the CiL and CoL traditions and the mere pragmatics of award justification in arbitration.

Moreover, if the parties agree beforehand that only the justifications provided by them are to be taken into consideration by the arbitrator, it does not appear that the resulting decision system should be *a priori* regarded as unfair. It should also be noted that foreseeability and effectiveness are also criteria for (procedural) justice.

---

[28] Compare I. A. Letia / A. Groza, *Structured argumentation in a mediator for Online Dispute Resolution*, in M. Baldoni et al (Eds.), *Declarative Agent Languages and Technologies V, 5th International Workshop, DALT 2007*, Honolulu, HI, USA, May 14, 2007, Springer, 2008, pp. 193-210.